%% file: main.tex
\documentclass[runningheads]{llncs}
\usepackage[T1]{fontenc}
\usepackage{graphicx}
\usepackage{url}
\usepackage{amsmath,amssymb}
\DeclareMathOperator*{\argmin}{arg\,min}
\usepackage{booktabs}
\usepackage{xcolor}

\begin{document}
\title{Predicting Brain Morphometry with MT-GNN: Mesh Evolution in Continuous Time with Graph-Based Metric Tensor Embeddings}

\titlerunning{Predicting Subcortical Surfaces via the Metric Tensor}
%
\author{Hao Ding\inst{1}, Daniel Semchin\inst{1}, Paul M. Thompson\inst{2} \and  Boris Gutman\inst{1}}
\authorrunning{H. Ding et al.}
\institute{Illinois Institute of Technology \\ \email{hding9@hawk.illinoistech.edu} \and
University of Southern California}

\maketitle
\begin{abstract}
\input{sections/00_abstract}

\keywords{Longitudinal shape prediction \and Metric tensor \and Differentiable reconstruction \and Realizability \and Continuous time \and ADNI.}
\end{abstract}
\input{sections/01_introduction}
\input{sections/02_related_work}
\input{sections/03_method}
\input{sections/04_experiments}

\input{sections/05_discussion}
\input{sections/06_conclusion}
\input{sections/07_acknowledgements}

\bibliographystyle{splncs04}
\bibliography{references}

\end{document}

%% file: sections/00_abstract.tex
Predicting how a subcortical structure's shape will evolve from a few prior scans could support prognosis and clinical-trial enrichment. Existing longitudinal mesh predictors either extrapolate shape trajectories via high-dimensional embeddings or regress vertex deformations directly. We instead predict the surface's intrinsic geometry in continuous time: a single per-structure graph network predicts the future per-vertex first fundamental form (metric tensor) for an arbitrary causal multiple-visit history and an arbitrary prediction horizon, conditioned on a Fourier encoding of the lead time. The predicted metric is decoded into a surface by a differentiable As-Rigid-As-Possible solver, and the model is trained end-to-end on the rigid-aligned vertex error. Training through the reconstruction keeps the decoded prediction a valid surface and consistently improves it. On 14 subcortical structures from the ADNI dataset, the proposed mesh evolution model (MT-GNN) predicts best among the evaluated methods at every horizon ($-2.29\%$ mean vertex error vs.\ the temporal mean, $p{=}6.1{\times}10^{-5}$, beating it on 14/14 structures), ahead of geodesic shape regression (DCM, $-0.19\%$) and a mesh transformer (TransforMesh, $-0.45\%$; $p{=}1.2{\times}10^{-4}$), with the lead widening as the horizon grows.

%% file: sections/01_introduction.tex
\section{Introduction}
\label{sec:intro}

Subcortical structures such as the hippocampus and amygdala atrophy measurably as
neurodegeneration progresses, and the shape of that atrophy carries information beyond
volume alone. Predicting how a structure's surface will look one to several years ahead, from a
short series of prior scans, could sharpen prognosis and enrich clinical trials with the subjects
most likely to progress. The difficulty is that the true per-visit shape change is small: after
rigid alignment it sits near the reproducibility floor of the segmentation and surfacing
pipeline. A predictor that simply averages the observed visits reproduces a denoised
``current'' shape instead of predicting change. This temporal mean is therefore a deceptively
strong baseline that is difficult to beat.

Existing longitudinal mesh predictors take one of two routes, and both struggle against this
baseline. The first regresses a trajectory in a shape space and extrapolates it forward,
e.g.\ by geodesic or hierarchical Riemannian regression on registered meshes
\cite{turkseven2023predicting}; this extrinsic extrapolation overshoots at long horizons. The
second learns to deform vertices directly with a deep network \cite{sarasua2021transformesh},
which can spend capacity modeling visit noise rather than the small longitudinal signal.
Both predict shape in extrinsic, ambient-space coordinates. We instead predict an intrinsic
descriptor: in our fixed-topology, registered setting, the surface's first fundamental form
(metric tensor) is an underused, rigid-frame-invariant summary of shape
\cite{gutman2015riemannian}. Predicting the metric, however, brings its own difficulty: supervising
it directly in metric space underperforms, for a reason specific to metric fields that we return
to below.

Our method, MT-GNN, combines two components.
The first is continuous-time intrinsic-metric prediction. A single lead-time-conditioned graph
network predicts the future first fundamental form from a causal multi-visit prefix at a
continuously parameterized lead time. The prediction is built in three stages: a log-Euclidean mean
of the input visits (the base metric), a shift of that mean along a learned population-displacement
direction (the base shift), and a per-vertex subject residual that corrects for individual
deviation. An as-rigid-as-possible (ARAP) reconstruction then decodes the predicted metric into a
surface.

The second is reconstruction-constrained training. A freely predicted metric field need not be
realizable by any embedded mesh. Our decoder, however, reads the field only through the edge lengths
it induces, and the ARAP solve returns the embedded mesh that best fits those lengths. We therefore
supervise the decoded surface rather than the metric itself, unlike a direct intrinsic loss on the
metric field (Ebin~\cite{ebin1970manifold} or affine-invariant). Because the reconstruction cannot
match lengths that no realizable surface admits, the vertex loss discourages such predictions. We
read this as a soft, empirical realizability constraint, not an exact projection onto the
realizable-metric manifold, and it consistently improves on a metric-space loss ($-2.26\%$ vs.\
$-1.78\%$ mean vertex error on the test set, \S\ref{sec:exp}).

\begin{figure}[htbp]
\centering
\includegraphics[width=\textwidth,trim=21 138 91 72,clip]{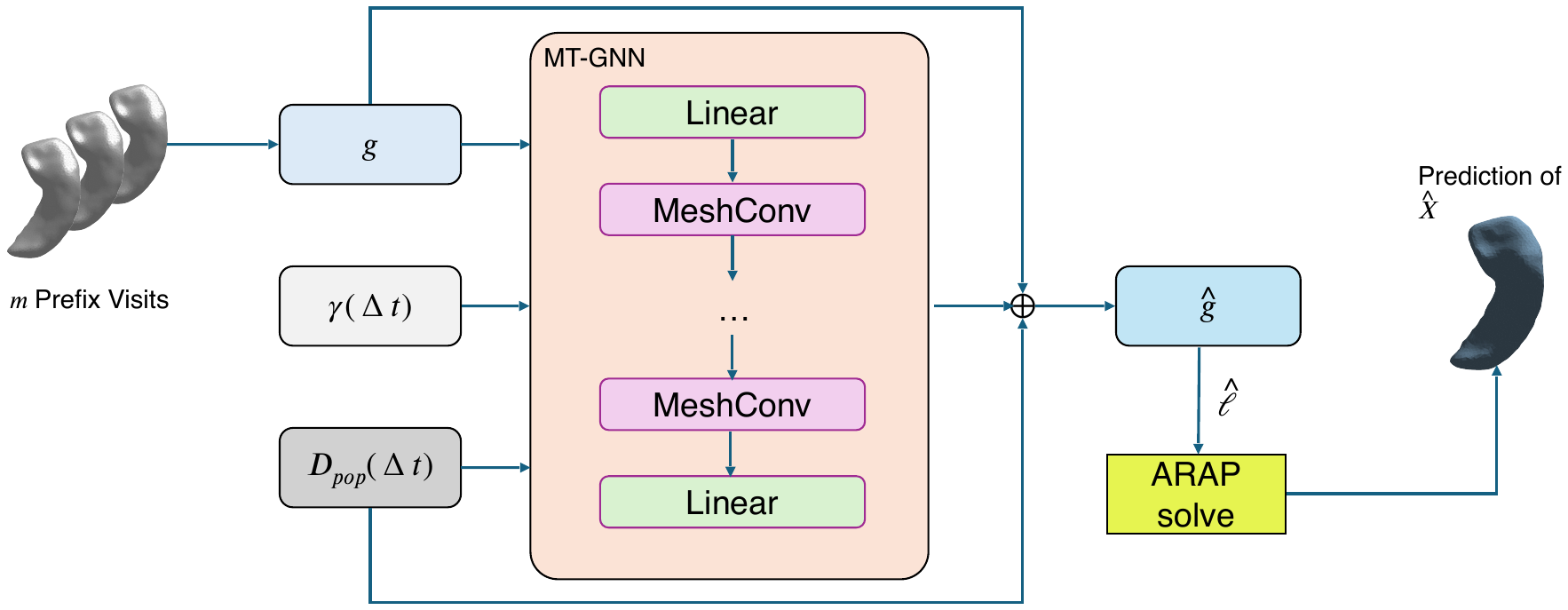}
\caption{\textbf{MT-GNN pipeline.} From an $m$-visit causal prefix, the per-vertex input metric
$g$, a Fourier time embedding $\gamma(\Delta t)$ of the lead time, and the population-displacement
term $D_{\mathrm{pop}}(\Delta t)$ feed a mesh graph network (alternating Linear and MeshConv
layers) that predicts a per-vertex log-metric residual. Combined ($\oplus$) with the log-Euclidean
base metric and the base shift $\beta\,D_{\mathrm{pop}}(\Delta t)$ (skip connections), this yields the
predicted first fundamental form $\hat g$, whose target edge lengths $\hat\ell$ are decoded into the
forecast surface $\widehat{\mathbf X}$ by a differentiable As-Rigid-As-Possible (ARAP) solve.}
\label{fig:pipeline}
\end{figure}


\noindent Our contributions can be summarized as follows:

\begin{enumerate}

  \item MT-GNN, a single continuous-time predictor of the intrinsic metric tensor
    whose prediction decomposes as log-Euclidean mean base $\rightarrow$ learned
    base shift $\rightarrow$ per-vertex subject residual, decoded through a differentiable
    ARAP solve, from a causal multi-visit prefix at a continuously parameterized lead time
    (Fig.~\ref{fig:pipeline}; \S\ref{sec:method}).
  \item Across 14 subcortical structures, MT-GNN improves on the temporal mean on all 14 and on both
    learned baselines, DCM \cite{turkseven2023predicting} and TransforMesh
    \cite{sarasua2021transformesh}, at every horizon, with the lead widening as the horizon grows;
    paired per-subject $t$-tests and per-term ablations support the comparison (\S\ref{sec:exp}).
  \item A design insight: a freely predicted metric field is generally not
    surface-realizable, and training through the differentiable reconstruction (rather
    than with a metric-space loss on the metric) acts as a soft realizability constraint and
    consistently improves accuracy (\S\ref{sec:exp:ablation}).
\end{enumerate}


%% file: sections/02_related_work.tex
\section{Related Work}
\label{sec:related}

A long line of work models anatomical change as a trajectory in a shape space and regresses it
from longitudinal observations. Geodesic and hierarchical Riemannian regression fit a smooth path
on a manifold of shapes and extrapolate it; the differential-coordinates formulation of T\"urkseven
et al.\ \cite{turkseven2023predicting} is a representative predictor and our primary trajectory
baseline (DCM). These methods excel when change is large and approximately geodesic, but their
extrinsic extrapolation overshoots once the assumed trajectory outruns the true, decelerating
change, an effect we observe at long horizons. We instead predict an intrinsic descriptor
of the target surface, referenced to a Riemannian mean of the inputs plus a population trend rather
than an extrapolated per-subject velocity.

On fixed-topology meshes, spiral and graph convolutions provide the efficient local operators our
predictor builds on \cite{gong2019spiralnet}, while graph-network simulators roll out mesh dynamics
\cite{pfaff2021meshgraphnets}. For longitudinal anatomy, TransforMesh \cite{sarasua2021transformesh}
applies a transformer over registered hippocampus meshes and is our learned-deformation baseline.
A complementary family predicts a per-element intrinsic quantity and integrates it: Neural Jacobian
Fields (NJF) \cite{aigerman2022njf} predict per-element deformation Jacobians.
We share NJF's ``predict-a-field-then-solve'' philosophy, but the contrast is one of
representation: NJF predicts extrinsic deformation Jacobians and reconstructs vertices
through a Poisson solver, which itself enforces integrability; we predict the
intrinsic metric tensor, where we find direct supervision using a tensor-native affine-invariant loss fails. This motivates
reconstruction-constrained training. For temporal conditioning we use Fourier features
\cite{tancik2020fourier} of the lead time, in the spirit of transformer positional encodings
\cite{vaswani2017attention}; this conditions the model directly on the horizon rather than
integrating a latent trajectory ODE \cite{rubanova2019latentode}.

Representing a closed genus-zero surface by its first and second fundamental forms underlies
intrinsic shape comparison \cite{gutman2015riemannian}; a freely-predicted metric field, however,
need not be exactly realizable by any embedded surface. Recovering vertices from
intrinsic data is classical: gradient-domain / Poisson editing \cite{yu2004poisson} and
As-Rigid-As-Possible (ARAP) modeling \cite{sorkine2007arap}, which we adopt as a differentiable decoder.
Because the metric is a field of symmetric-positive-definite matrices, we process it with
established SPD geometry. We use log-Euclidean coordinates \cite{arsigny2006logeuclidean} for the base tensor field and the individual subject residual and a log-Cholesky parameterization \cite{lin2019cholesky} to keep
redictions SPD. We compare the ARAP objective with a loss based on the affine-invariant metric \cite{pennec2006riemannian}, a loss intrinsic to metric tensor fields but with no realizability enforcement.

%% file: sections/03_method.tex
\section{Method}
\label{sec:method}

\subsection{Problem Formulation}
\label{sec:method:setup}
For a given structure, all subjects' scans share the same registered~\cite{GutmanShapeRegistration}, fixed-topology genus-zero triangle mesh
$\mathcal{M}=(\mathcal{V},\mathcal{F})$, where $\mathcal{V}$ is the vertex set with $N=|\mathcal{V}|$
vertices, $\mathcal{F}$ is the set of triangular faces, and $\mathcal{E}$ is the induced edge set.
The subjects also share one spherical parameterization, which assigns each vertex $i$ a fixed tangent
frame. The faces $\mathcal{F}$ fix the connectivity that the cotangent Laplacian uses later in the
reconstruction of \S\ref{sec:method:recon}. A \emph{visit} of a subject is an embedding of
$\mathcal{M}$. It gives the vertex positions $\mathbf{X}\in\mathbb{R}^{N\times3}$ and the induced per-vertex
intrinsic geometry: the first fundamental form (metric tensor) $g_i\in\mathrm{SPD}(2)$ and the mean
curvature $H_i\in\mathbb{R}$. Both are expressed in the fixed tangent frame.

A subject contributes a longitudinal sequence of visits at ages (in months) $t_1<t_2<\dots$. Given
an arbitrary causal $m$-visit prefix $\mathcal{P}=\big(\mathbf{X}^{(k)},g^{(k)},H^{(k)}\big)_{k=1}^{m}$
at times $t_1<t_2<\dots<t_m$ and a lead time $\Delta t>0$, the task is to predict the surface
$\mathbf{X}^{\tau}$ at the future age $\tau=t_m+\Delta t$. We predict
$F_\theta:(\mathcal{P},\Delta t)\mapsto \widehat{\mathbf{X}}$ for arbitrary prefixes and
arbitrary $\Delta t$.

\subsubsection{Evaluation metric.}
We score a prediction $\widehat{\mathbf{X}}$ against the
ground-truth $\mathbf{X}^{\tau}$ by the rigid motion-invariant \emph{mean vertex error} (MVE),
\begin{equation}
\begin{aligned}
  \mathrm{MVE}(\widehat{\mathbf{X}}, \mathbf{X}^{\tau})
    &\;=\; \frac{1}{N}\sum_{i=1}^{N}\big\| \mathbf{R}^{\star}\widehat{\mathbf{x}}_i + \mathbf{t}^{\star} - \mathbf{x}^{\tau}_i \big\|_2, \\
  (\mathbf{R}^{\star},\mathbf{t}^{\star}) &\;=\; \argmin_{\mathbf{R}\in SO(3),\,\mathbf{t}\in\mathbb{R}^3}\ \sum_{i=1}^{N}\big\| \mathbf{R}\,\widehat{\mathbf{x}}_i + \mathbf{t} - \mathbf{x}^{\tau}_i \big\|_2^2,
\end{aligned}
  \label{eq:mve}
\end{equation}
where $\widehat{\mathbf{x}}_i,\mathbf{x}^{\tau}_i\in\mathbb{R}^3$ are the $i$-th vertices of $\widehat{\mathbf{X}},\mathbf{X}^{\tau}$, and $(\mathbf{R}^{\star},\mathbf{t}^{\star})$ is the least-squares rigid Procrustes (Kabsch) alignment: rotation only ($SO(3)$, no reflection) plus translation, no scale.
We use the same measure of prediction quality during training and evaluation.

\subsection{Surface and Metric Representation}
\label{sec:method:repr}
The metric tensor is a field of $2\times2$ symmetric-positive-definite matrices, and we process it
in the log-domain of $\mathrm{SPD}(2)$. The matrix logarithm $\mathrm{Log}$ sends $\mathrm{SPD}(2)$ to
the real symmetric matrices $\mathrm{Sym}(2)$. Because $\mathrm{Sym}(2)$ is diffeomorphic to $\mathbb{R}^3$, we use this for an unconstrained parameterization: any predicted $\ell\in\mathbb{R}^3$ maps to a valid metric
$\mathrm{Exp}\,\ell\in\mathrm{SPD}(2)$ \cite{arsigny2006logeuclidean,lin2019cholesky}.

\subsubsection{Base metric.}
We use the log-Euclidean mean of the prefix metric tensors,
\begin{equation}
  K_i \;=\; \tfrac{1}{m}\sum_{k=1}^{m}\mathrm{Log}\,g^{(k)}_i \in \mathbb{R}^3,
  \qquad \bar g_i = \mathrm{Exp}\,K_i,
  \label{eq:base}
\end{equation}
a closed-form, swelling-free SPD mean as a practical surrogate for the
affine-invariant Karcher mean \cite{pennec2006riemannian}.

\subsubsection{Temporal-mean surface.}
The reconstruction (\S\ref{sec:method:recon}) needs an initial embedding. We rigidly align the
later prefix visits to the first and average,
\begin{equation}
\begin{aligned}
  \mathbf{s}^0_i &\;=\; \tfrac{1}{m}\Big(\mathbf{x}^{(1)}_i + \sum_{k=2}^{m}\big(\mathbf{R}_k\,\mathbf{x}^{(k)}_i+\mathbf{t}_k\big)\Big), \\
  (\mathbf{R}_k,\mathbf{t}_k) &\;=\; \argmin_{\mathbf{R}\in SO(3),\,\mathbf{t}\in\mathbb{R}^3}\sum_i\|\mathbf{R}\,\mathbf{x}^{(k)}_i+\mathbf{t}-\mathbf{x}^{(1)}_i\|_2^2,
\end{aligned}
  \label{eq:anchor}
\end{equation}
where $\mathbf{s}^0_i$ is the $i$-th vertex of the temporal-mean surface $\mathbf{S}^0$, which is exactly the
temporal mean used as the reference baseline in \S\ref{sec:exp}.

\subsubsection{Population displacement.}
One signal that the subject's own history cannot provide is how the population changes on average at
a given lead time. We call this the population displacement. From the training set we estimate,
for each horizon $\Delta t$, the mean log-domain change of the future-visit metric $g^{\tau}$ relative
to the base $K$,
\begin{equation}
  D_i(\Delta t) \;=\; \operatorname*{mean}_{\mathcal{P}\,:\,\text{lead}=\Delta t}
     \big(\mathrm{Log}\,g^{\tau}_i - K_i\big) \in \mathbb{R}^3,
  \label{eq:dpop}
\end{equation}
computed once and frozen (no test-time leakage; $D$ depends only on training targets).

\subsubsection{Time encoding and per-vertex features.}
We encode the lead time with a Fourier feature map $\gamma$, in the spirit of positional encodings
\cite{tancik2020fourier,vaswani2017attention}. For a fixed set of frequencies $\{\omega_k\}$,
\begin{equation}
  \gamma(\Delta t) \;=\; \big(\,\sin(\omega_k\,\Delta t),\ \cos(\omega_k\,\Delta t)\,\big)_{k}.
  \label{eq:fourier}
\end{equation}
The per-vertex input feature $\phi_i$ concatenates the base metric, the $m$ prefix log-metrics, the
time encoding $\gamma(\Delta t)$ (shared across vertices), and the population displacement,
\begin{equation}
  \phi_i \;=\; \big[\,K_i,\ \{\mathrm{Log}\,g^{(k)}_i\}_{k=1}^{m},\ \gamma(\Delta t),\ D_i(\Delta t)\,\big].
  \label{eq:feat}
\end{equation}

\subsection{Metric Prediction}
\label{sec:method:forecast}
A graph network $f_\theta$ \cite{gong2019spiralnet} over adjacency of $\mathcal{M}$ maps the per-vertex features
$\Phi=\{\phi_i\}$ to a per-vertex log-metric subject residual $r_i=f_\theta(\Phi)_i\in\mathbb{R}^3$. In our setting, $f_\theta$ has four mesh graph-convolution layers, with cotangent normalization of the graph. The predicted metric is the sum of three interpretable terms,
\begin{equation}
  \mathrm{Log}\,\hat g_i \;=\; \underbrace{K_i}_{\text{base metric}}
     \;+\; \underbrace{\beta\,D_i(\Delta t)}_{\text{base shift}}
     \;+\; \underbrace{r_i}_{\text{subject residual}},
  \label{eq:model}
\end{equation}
where $\beta\in\mathbb{R}$ is a learnable base-shift scalar. The final layer of $f_\theta$ is
zero-initialized, and training begins exactly at the base metric $+$ base shift, with the network only learning
per-vertex deviations. We deliberately do not extrapolate a per-subject temporal slope: at
this noise level the individual slope is not a reliable direction (\S\ref{sec:exp:ablation}), and the
only extrapolation is along the population trend $D(\Delta t)$. The $\mathrm{Exp}$ map keeps
$\hat g_i\in\mathrm{SPD}(2)$ by construction.

\subsection{Differentiable Reconstruction}
\label{sec:method:recon}
We decode the predicted metric into a surface with a robust As-Rigid-As-Possible (ARAP) local--global
solver \cite{sorkine2007arap}. Each edge $(i,j)\in\mathcal{E}$ has a fixed tangent-plane offset
$\mathbf{e}_{ij}$ from the spherical parameterization; the predicted metric sets its target squared length by
symmetrizing over the two endpoints,
\begin{equation}
  \hat\ell^{\,2}_{ij} \;=\; \tfrac12\big(\mathbf{e}_{ij}^{\top}\hat g_i\,\mathbf{e}_{ij} + \mathbf{e}_{ij}^{\top}\hat g_j\,\mathbf{e}_{ij}\big).
  \label{eq:edgelen}
\end{equation}
Given target edge lengths, ARAP seeks vertex positions whose edge geometry is as-rigid-as-possible
consistent with $\hat\ell$,
\begin{equation}
  \widehat{\mathbf{X}} \;=\; \argmin_{\mathbf{X}}\ \sum_{i}\sum_{j\in\mathcal{N}(i)} w_{ij}\,
     \big\| (\mathbf{x}_i-\mathbf{x}_j) - \mathbf{R}_i\,\hat\ell_{ij}\,\hat{\mathbf{u}}_{ij} \big\|_2^2,
  \label{eq:arap}
\end{equation}
Here $\mathcal{N}(i)$ is the one-ring neighborhood of vertex $i$, $w_{ij}$ are cotangent weights,
$\mathbf{R}_i\in SO(3)$ are per-vertex rotations, and $\hat{\mathbf{u}}_{ij}$ is the unit rest-edge direction taken from
the temporal-mean surface $\mathbf{S}^0$. Only the target length $\hat\ell_{ij}$ depends on the predicted
metric $\hat g$; optimizing Eq.~\eqref{eq:arap} with respect to $\hat g$ is a straightforward application of the chain rule. We solve Eq.~\eqref{eq:arap} by the standard local--global alternation
initialized at the temporal-mean surface $\mathbf{S}^0$: a local step fixes $\mathbf{X}$ and updates each $\mathbf{R}_i$ by a small SVD,
and a global step fixes $\{\mathbf{R}_i\}$ and solves the linear system
$(\mathbf{L}+\lambda \mathbf{I})\,\mathbf{X} = \mathbf{b}(\{\mathbf{R}_i\})$, where $\mathbf{L}$ is the cotangent Laplacian and $\lambda>0$ is a small
regularizer that keeps the system positive-definite (the Laplacian alone is singular on a global
translation). Because $\mathbf{L}$ depends only on the fixed connectivity, its Cholesky factor is precomputed
once and shared across all iterations and all samples. We unroll a fixed number of $15$ iterations, making
$\widehat{\mathbf{X}} = \mathrm{ARAP}(\mathbf{S}^0,\hat g)$ differentiable end-to-end in $\hat g$. This is the
gradient-domain principle of Poisson surface editing \cite{yu2004poisson}, specialized to an
intrinsic edge-length target.

\subsection{Training Objective}
\label{sec:method:train}
The predictor is trained by minimizing the rigid-aligned vertex error Eq.~\eqref{eq:mve} of the
reconstructed surface, backpropagated through the unrolled solver. Over a training set of
prefix--target pairs $\{\mathcal{P}\}$ we minimize
\begin{equation}
  \mathcal{L}(\theta,\beta)
   \;=\;\frac{\sum_{\mathcal{P}} w_{\mathcal{P}}\,\mathrm{MVE}\big(\mathrm{ARAP}(\mathbf{S}^0,\hat g),\,\mathbf{X}^{\tau}\big)}
             {\sum_{\mathcal{P}} w_{\mathcal{P}}}
   \;+\; \eta\,\frac{1}{N}\sum_{i}\|r_i\|_2^2 ,
  \label{eq:loss}
\end{equation}
where $\hat g$ depends on $(\theta,\beta)$ through Eq.~\eqref{eq:model}, $\eta$ weights a small subject-residual regularizer, and $w_{\mathcal{P}}=w_{\mathrm{subj}}\!\cdot w_{\Delta t}$ balances per-subject
contributions and the (heavily skewed) horizon distribution. Because the loss is computed on the
reconstructed vertices rather than on $\hat g$ directly, it shapes $\hat g$ only through the surface
it reconstructs using Eq.~\eqref{eq:arap}, which is always a valid embedded mesh. This is a soft realizability constraint: approximate and reconstruction-defined, not an exact
Gauss--Codazzi projection. 

\subsection{Metric Tensor plus Mean Curvature  Variant (MT-GNN+H)}
\label{sec:method:gnnh}
To test whether mean curvature $H$ adds anything once $g$ is predicted, MT-GNN+H reuses the graph backbone of
Eq.~\eqref{eq:model} and attaches a second zero-initialized head. Beside the log-metric residual
$r_i$, an $H$-head predicts a bounded mean-curvature residual on the prefix-mean curvature
$\bar H_i=\tfrac{1}{m}\sum_{k=1}^{m}H^{(k)}_i$,
\begin{equation}
  \hat H_i \;=\; \bar H_i \;+\; \tanh(h_i)\,\kappa\,\sigma_H ,
  \label{eq:hhead}
\end{equation}
where $h_i$ is the head output, $\sigma_H$ is the curvature scale over the training set, and $\kappa$
bounds the step; the zero initialization starts prediction at $\bar H_i$, mirroring how the metric
starts at the base metric.

The predicted metric and curvature are decoded jointly. We keep the differentiable ARAP objective of
Eq.~\eqref{eq:arap} and add one curvature term to its global step. The discrete mean-curvature identity defines the cotangent Laplacian of the vertex positions as $\mathbf{L}\mathbf{X}=\mathbf{c}$
with per-vertex rows $\mathbf{c}_i=2A_i\hat H_i\,\mathbf{n}_i$ (twice the vertex area times mean
curvature along the surface normal). Enforcing it softly, the global step becomes
\begin{equation}
  \big(\mathbf{L}+\lambda\mathbf{I}+\mu\,\mathbf{L}^{\!\top}\mathbf{L}\big)\,\widehat{\mathbf{X}}
    \;=\; \mathbf{b}(\{\mathbf{R}_i\}) \;+\; \mu\,\mathbf{L}^{\!\top}\mathbf{c},
  \label{eq:araph}
\end{equation}
where $\mu\ge0$ weights the curvature term. The normals $\mathbf{n}_i$ and areas $A_i$ are frozen at
the temporal-mean surface $\mathbf{S}^0$, which makes $\mathbf{c}$ is a fixed right-hand side rather than a
moving target; this keeps the combined optimization stable and its Cholesky factor precomputable, and
leaves the result differentiable in $\hat H$. The remaining aspects of the problem, including the features, the training objective
Eq.~\eqref{eq:loss}, and validation-only selection are unchanged; MT-GNN+H differs from
MT-GNN by the curvature head alone, and $\mu=0$ recovers the $g$-only prediction through the same solver.

%% file: sections/04_experiments.tex
\section{Experiments}
\label{sec:exp}

\subsection{Data, Protocol, and Baselines}
\label{sec:exp:data}
We use longitudinal T1 MRI from ADNI \cite{jack2008adni} processed with FreeSurfer~8 (v8.0.0)
\cite{fischl2012freesurfer}, taking the 14 subcortical structures (left/right thalamus, caudate,
putamen, pallidum, hippocampus, amygdala, accumbens). Each visit gives a registered~\cite{GutmanShapeRegistration},
fixed-topology genus-zero mesh with per-vertex metric and curvature; a per-sample outlier flag based on Mean Absolute Distance and the persistence error removes segmentation and surfacing failures. From the
quality-controlled cohort we build \emph{causal pairs}: arbitrary causal three-visit prefixes paired with a future target at lead time
$\Delta t\in\{12,24,36,48\}$ months. We use a subject-disjoint split (train 405 / val 86
/ test 86) and report all results below on the test set. Errors are in
the $\approx0.5$--$0.9$\,mm range; we report them as percentage change relative to the
temporal mean, a subject-internal rigidly-aligned average of the input visits, which is
strong because per-visit change is near the noise floor. We train MT-GNN per structure with Adam (learning rate $10^{-3}$, batch $16$, gradient-norm clip
$1.0$), a cosine schedule over $150$ epochs, and a graph network of hidden width $128$.

We compare against two learned baselines. The first, DCM, is the geodesic/hierarchical Riemannian regressor
\cite{turkseven2023predicting}, adopted to continuous lead times by extrapolating a population-level velocity; the second,
TransforMesh (TfM), is the mesh transformer \cite{sarasua2021transformesh} retrofitted with
Fourier-time conditioning and temporal-mean-referenced. Both baselines are adapted to the arbitrary-prefix /
arbitrary-horizon protocol and evaluated on the identical causal test pairs, metric, and temporal-mean reference as our own MT-GNN. 

\subsection{Main Results}
\label{sec:exp:main}
\begin{figure}[htbp]
\centering
\includegraphics[width=0.72\linewidth]{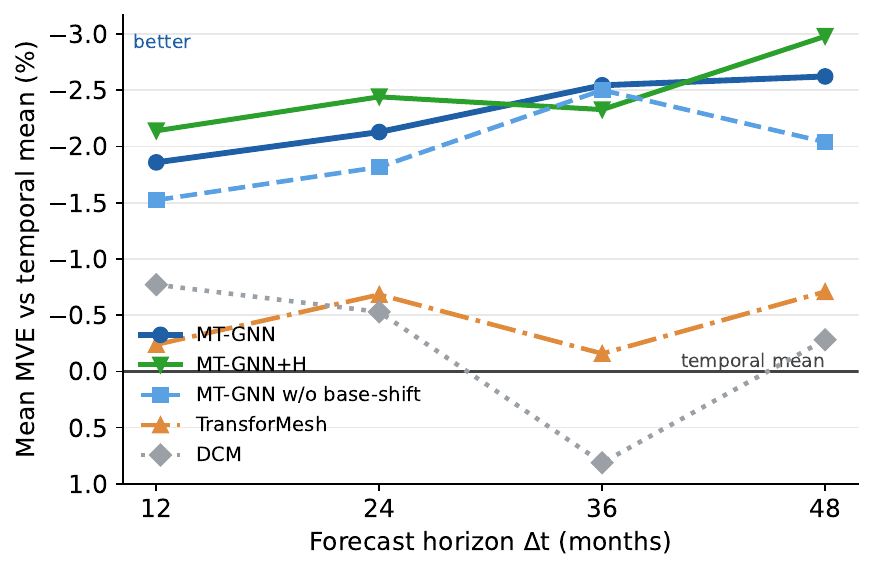}
\caption{\textbf{Accuracy vs.\ horizon (14 structures).} Mean MVE relative to
the temporal mean (\%, lower is better; $y$-axis inverted) at each lead time $\Delta t$. MT-GNN
and MT-GNN+H are the top two lowest-error curves, with MT-GNN improving monotonically and its lead over
the temporal mean widening with $\Delta t$, while DCM rises above the temporal mean at $\Delta t{=}36$; the MT-GNN+H gain over
MT-GNN is small and significant per structure in only $4/14$ structures (Table~\ref{tab:pvalues}).}
\label{fig:horizon}
\end{figure}

Table~\ref{tab:main} and Fig.~\ref{fig:horizon} report mean MVE (mm) at each horizon.
MT-GNN and its mean curvature variant MT-GNN+H have the two best horizon-averaged errors (mean $-2.29\%$ and $-2.48\%$), ahead of MT-GNN w/o base-shift ($-1.97\%$),
TransforMesh ($-0.45\%$), and DCM ($-0.19\%$). MT-GNN is lower than all three at every horizon and its
lead over the temporal mean grows with horizon; MT-GNN+H is lowest at $\Delta t{=}12,24,48$ but slips
just behind the no-shift ablation at $\Delta t{=}36$. Both beat the
temporal mean on all structures (Table~\ref{tab:perstruct}), significantly so in
$12/14$ (MT-GNN) and $14/14$ (MT-GNN+H) by paired $t$-tests across subjects after Bonferroni
correction over the five methods $\times$ $14$ structures ($p<0.05/70$). Against the learned baselines (Table~\ref{tab:pvalues}), MT-GNN and
MT-GNN+H significantly outperform TransforMesh in $11$--$12/14$ structures and DCM in $8$--$9/14$.

\begin{table}[htbp]
\caption{Mean vertex error (MVE) in mm on the test dataset at each horizon
(per-structure median over pairs, averaged over the 14 structures), the horizon-averaged mean, and
\% change vs.\ the temporal mean (negative is better).
}
\label{tab:main}
\centering
\setlength{\tabcolsep}{5pt}
\resizebox{\textwidth}{!}{%
\begin{tabular}{lcccccc}
\toprule
Method & $\Delta t$12 & $\Delta t$24 & $\Delta t$36 & $\Delta t$48 & Mean & $\Delta\%$ \\
\midrule
Temporal mean (ref.) & 0.6184 & 0.6203 & 0.6355 & 0.6522 & 0.6316 & --- \\
DCM \cite{turkseven2023predicting} & 0.6137 & 0.6170 & 0.6407 & 0.6504 & 0.6304 & $-0.19$ \\
TransforMesh \cite{sarasua2021transformesh} & 0.6170 & 0.6160 & 0.6345 & 0.6476 & 0.6288 & $-0.45$ \\
MT-GNN w/o base-shift & 0.6090 & 0.6090 & 0.6196 & 0.6390 & 0.6191 & $-1.97$ \\
\textbf{MT-GNN} & $\mathbf{0.6070}$ & $\mathbf{0.6071}$ & $\mathbf{0.6193}$ & $\mathbf{0.6352}$ & $\mathbf{0.6171}$ & $\mathbf{-2.29}$ \\
\textbf{MT-GNN+H} & $\mathbf{0.6052}$ & $\mathbf{0.6051}$ & $\mathbf{0.6207}$ & $\mathbf{0.6328}$ & $\mathbf{0.6160}$ & $\mathbf{-2.48}$ \\
\bottomrule
\end{tabular}}
\end{table}

\begin{figure}[htbp]
\centering
\includegraphics[width=0.72\linewidth]{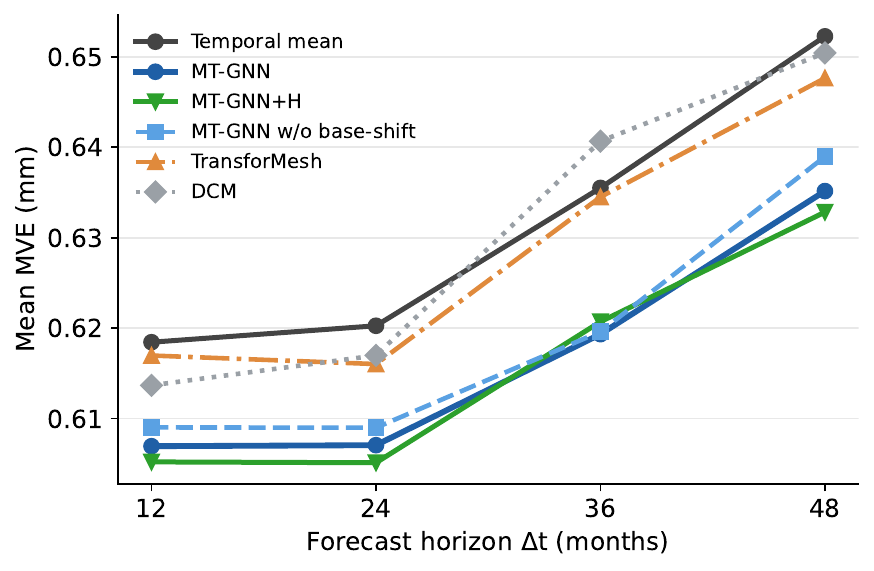}
\caption{\textbf{Absolute accuracy vs.\ horizon (14 structures).} Mean MVE in
millimetres at each horizon. Errors lie in a narrow band near the segmentation/surfacing
reproducibility floor; MT-GNN and MT-GNN+H are the lowest-error curves and separate further
from the temporal mean as the horizon grows, with a small gap between them.}
\label{fig:abshorizon}
\end{figure}

Figure~\ref{fig:abshorizon} shows the same comparison in absolute terms. All methods sit within a
narrow band ($\approx0.05$ \,mm) near the FreeSurfer~8 reproducibility floor, confirming that
the per-visit change is small; MT-GNN's gap to the temporal mean, though sub-milimetre, widens with horizon.

\subsection{Per-Structure Results}
\label{sec:exp:perstruct}
Table~\ref{tab:perstruct} breaks the results down by structure. On 11 of the 14 structures, the best of the five methods is MT-GNN or MT-GNN+H. The three exceptions do not reach statistical significance on left caudate,
where TransforMesh is best, and left and right accumbens, where the no-shift ablation edges ahead.
The largest improvement is on the hippocampus (right hippocampus $-5.38\%$), with strong gains
also on the amygdala (left amygdala $-3.58\%$); both are structures whose atrophy is comparatively
stereotyped, matching the horizon trend of \S\ref{sec:exp:main}. DCM is the one method whose aggregate changes sign between the two tables ($-0.19\%$ in Table~\ref{tab:main}, $+0.11\%$ here): its gains are
concentrated on high-error structures such as the hippocampus, which dominate Table~\ref{tab:main}'s
millimetre-pooled ratio but count as single equal-weighted entries in this table's across-structure
mean of per-structure percentages.

\begin{table}[htbp]
\caption{\textbf{Per-structure results.} MVE \% vs.\ the temporal mean
(negative is better), for each structure averaged over all test pairs across all horizons
$\Delta t\in\{12,24,36,48\}$ months.
because short horizons are more numerous, this pooled average weights them
more heavily. The Mean row is an unweighted across-structure mean of the per-structure
percentages, which differs from the pooled aggregate of Table~\ref{tab:main} (see text).
}
\label{tab:perstruct}
\centering
\small
\setlength{\tabcolsep}{6pt}
\resizebox{\textwidth}{!}{%
\begin{tabular}{lccccc}
\toprule
Structure & MT-GNN & MT-GNN+H & \begin{tabular}[c]{@{}c@{}}MT-GNN\\w/o base-shift\end{tabular} & TransforMesh & DCM \\
\midrule
L-Thal & $-3.29$ & $\mathbf{-3.61}$ & $-3.21$ & $-1.59$ & $+0.56$ \\
L-Caud & $-0.69$ & $-2.02$ & $-0.26$ & $\mathbf{-2.26}$ & $-2.18$ \\
L-Put  & $\mathbf{-1.98}$ & $-1.30$ & $-0.66$ & $-0.39$ & $+1.34$ \\
L-Pall & $-1.86$ & $\mathbf{-2.21}$ & $-1.52$ & $-0.18$ & $-0.60$ \\
L-Hipp & $-2.22$ & $\mathbf{-2.95}$ & $-1.82$ & $-0.23$ & $-0.40$ \\
L-Amyg & $\mathbf{-3.58}$ & $-3.46$ & $-3.18$ & $-2.98$ & $+0.43$ \\
L-Accu & $-3.90$ & $-3.27$ & $\mathbf{-3.91}$ & $+0.60$ & $-0.02$ \\
R-Thal & $\mathbf{-2.67}$ & $-2.57$ & $-2.20$ & $-1.25$ & $-1.75$ \\
R-Caud & $\mathbf{-0.48}$ & $-0.17$ & $-0.04$ & $+0.75$ & $+0.81$ \\
R-Put  & $-1.91$ & $\mathbf{-1.98}$ & $-0.29$ & $+0.37$ & $+0.73$ \\
R-Pall & $-0.73$ & $\mathbf{-0.76}$ & $-0.29$ & $+0.12$ & $+2.39$ \\
R-Hipp & $\mathbf{-5.38}$ & $-5.05$ & $-4.92$ & $-1.91$ & $-3.39$ \\
R-Amyg & $-1.17$ & $\mathbf{-2.33}$ & $-1.68$ & $+1.72$ & $+3.04$ \\
R-Accu & $-1.87$ & $-1.76$ & $\mathbf{-1.96}$ & $+0.92$ & $+0.53$ \\
\midrule
Mean & $-2.26$ & $\mathbf{-2.39}$ & $-1.85$ & $-0.45$ & $+0.11$ \\
\bottomrule
\end{tabular}}
\end{table}

\begin{table}[htbp]
\caption{\textbf{Per-structure paired significance.} Paired $t$-test across subjects (one MVE per
subject per structure, aggregated over that subject's test pairs). Each cell gives the better method
(\textbf{G}$=$MT-GNN, \textbf{H}$=$MT-GNN+H, \textbf{D}$=$DCM, \textbf{T}$=$TransforMesh) and the
$p$-value; $^{*}$ marks significance after Bonferroni correction ($p<0.05/70\approx7.1\times10^{-4}$,
for $5$ comparisons $\times$ $14$ structures).}
\label{tab:pvalues}
\centering
\footnotesize
\setlength{\tabcolsep}{4pt}
\resizebox{\textwidth}{!}{%
\begin{tabular}{lccccc}
\toprule
Structure & MT-GNN vs +H & MT-GNN vs DCM & MT-GNN vs TfM & +H vs DCM & +H vs TfM \\
\midrule
L-Thal & H\,$6.1\mathrm{e}{-}5^{*}$ & G\,$2.5\mathrm{e}{-}3$ & G\,$1.5\mathrm{e}{-}4^{*}$ & H\,$4.4\mathrm{e}{-}5^{*}$ & H\,$2.4\mathrm{e}{-}7^{*}$ \\
L-Caud & H\,$6.8\mathrm{e}{-}6^{*}$ & D\,$1.8\mathrm{e}{-}2$ & T\,$4.7\mathrm{e}{-}2$ & D\,$4.5\mathrm{e}{-}1$ & T\,$9.1\mathrm{e}{-}1$ \\
L-Put  & H\,$2.8\mathrm{e}{-}5^{*}$ & G\,$5.1\mathrm{e}{-}2$ & G\,$1.1\mathrm{e}{-}8^{*}$ & H\,$2.5\mathrm{e}{-}3$ & H\,$2.4\mathrm{e}{-}13^{*}$ \\
L-Pall & H\,$3.1\mathrm{e}{-}5^{*}$ & G\,$1.1\mathrm{e}{-}8^{*}$ & G\,$1.7\mathrm{e}{-}7^{*}$ & H\,$1.6\mathrm{e}{-}10^{*}$ & H\,$4.7\mathrm{e}{-}11^{*}$ \\
L-Hipp & G\,$4.9\mathrm{e}{-}1$ & G\,$7.1\mathrm{e}{-}10^{*}$ & G\,$1.6\mathrm{e}{-}15^{*}$ & H\,$8.8\mathrm{e}{-}9^{*}$ & H\,$4.4\mathrm{e}{-}16^{*}$ \\
L-Amyg & H\,$7.2\mathrm{e}{-}3$ & G\,$7.9\mathrm{e}{-}8^{*}$ & G\,$9.1\mathrm{e}{-}9^{*}$ & H\,$7.5\mathrm{e}{-}10^{*}$ & H\,$3.4\mathrm{e}{-}9^{*}$ \\
L-Accu & H\,$3.9\mathrm{e}{-}2$ & G\,$3.2\mathrm{e}{-}5^{*}$ & G\,$3.3\mathrm{e}{-}9^{*}$ & H\,$7.9\mathrm{e}{-}6^{*}$ & H\,$7.4\mathrm{e}{-}9^{*}$ \\
R-Thal & H\,$6.6\mathrm{e}{-}1$ & G\,$2.4\mathrm{e}{-}3$ & G\,$2.0\mathrm{e}{-}22^{*}$ & H\,$1.7\mathrm{e}{-}3$ & H\,$8.5\mathrm{e}{-}17^{*}$ \\
R-Caud & H\,$2.9\mathrm{e}{-}3$ & D\,$9.0\mathrm{e}{-}1$ & G\,$2.1\mathrm{e}{-}1$ & H\,$5.6\mathrm{e}{-}1$ & H\,$1.7\mathrm{e}{-}3$ \\
R-Put  & H\,$6.2\mathrm{e}{-}1$ & G\,$3.3\mathrm{e}{-}1$ & G\,$1.0\mathrm{e}{-}3$ & H\,$3.2\mathrm{e}{-}1$ & H\,$2.3\mathrm{e}{-}6^{*}$ \\
R-Pall & H\,$1.5\mathrm{e}{-}3$ & G\,$2.6\mathrm{e}{-}5^{*}$ & G\,$3.9\mathrm{e}{-}4^{*}$ & H\,$4.9\mathrm{e}{-}9^{*}$ & H\,$1.7\mathrm{e}{-}8^{*}$ \\
R-Hipp & G\,$1.9\mathrm{e}{-}1$ & G\,$4.5\mathrm{e}{-}6^{*}$ & G\,$4.1\mathrm{e}{-}8^{*}$ & H\,$3.1\mathrm{e}{-}6^{*}$ & H\,$7.2\mathrm{e}{-}8^{*}$ \\
R-Amyg & H\,$1.0\mathrm{e}{-}1$ & G\,$2.7\mathrm{e}{-}4^{*}$ & G\,$1.2\mathrm{e}{-}7^{*}$ & H\,$1.9\mathrm{e}{-}4^{*}$ & H\,$1.5\mathrm{e}{-}12^{*}$ \\
R-Accu & H\,$2.5\mathrm{e}{-}2$ & G\,$5.2\mathrm{e}{-}11^{*}$ & G\,$9.0\mathrm{e}{-}11^{*}$ & H\,$2.2\mathrm{e}{-}12^{*}$ & H\,$3.4\mathrm{e}{-}12^{*}$ \\
\bottomrule
\end{tabular}}
\end{table}

\begin{figure}[htbp]
\centering
\includegraphics[width=\textwidth]{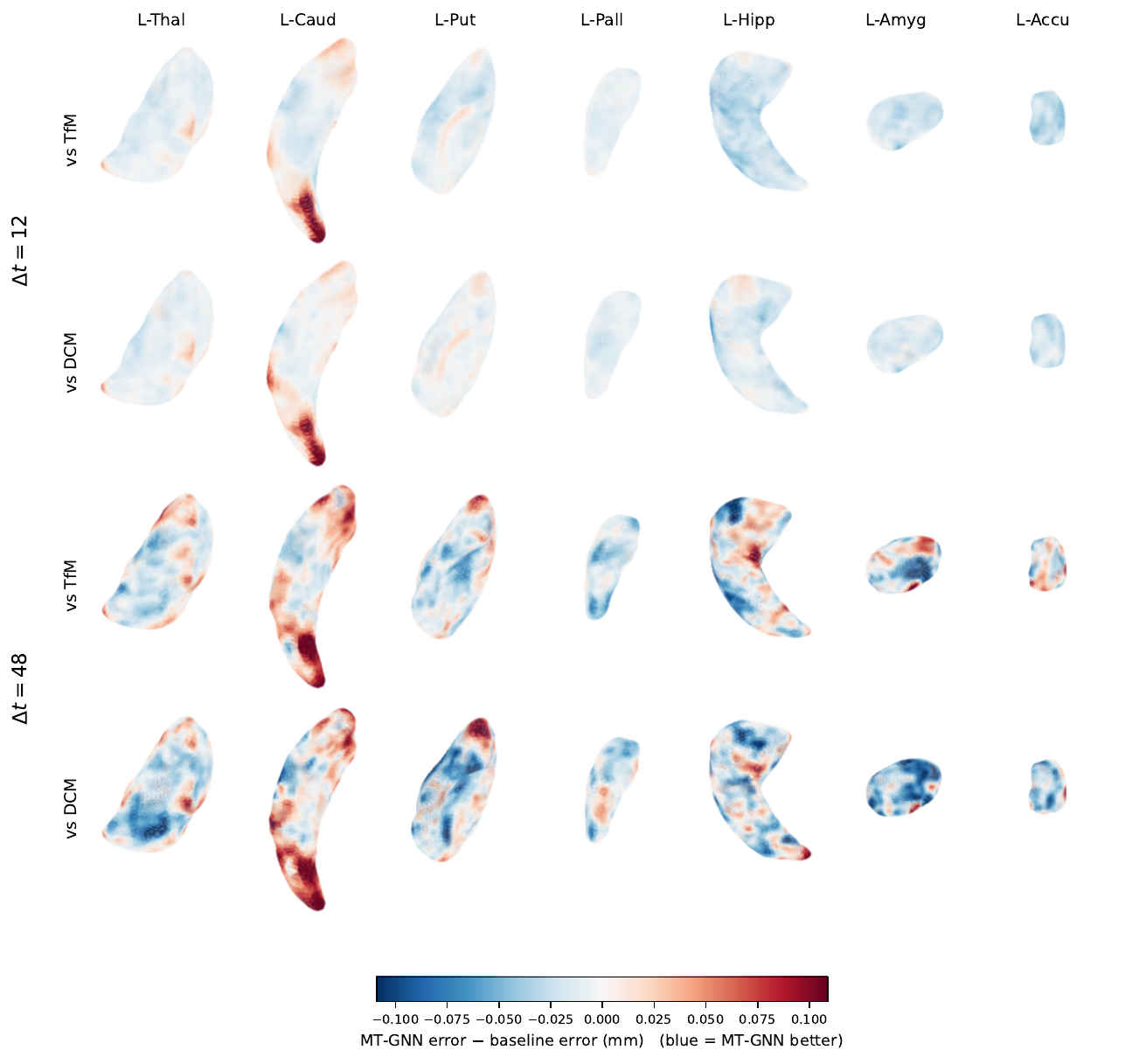}
\caption{\textbf{Per-structure error difference, averaged over the test set.} Per-vertex
MT-GNN error $-$ baseline error (mm) against TransforMesh and DCM at $\Delta t{=}12$ and $48$,
for the seven left structures (blue $=$ MT-GNN lower error, i.e.\ better). The advantage is small
at the near horizon and grows and localizes by the far horizon, most against DCM.}
\label{fig:deltamap}
\end{figure}

Figure~\ref{fig:deltamap} maps the per-vertex error difference (MT-GNN $-$ baseline), averaged over
all test pairs, for the seven left structures against TransforMesh and DCM at the near and far
horizons; blue marks vertices where MT-GNN is closer to the ground-truth surface. At $\Delta t{=}12$
the advantage is small and diffuse, in line with the near-floor errors of Fig.~\ref{fig:abshorizon};
by $\Delta t{=}48$ it is larger and spatially concentrated, and broadest against DCM whose
extrapolation overshoots, echoing the widening lead of Fig.~\ref{fig:horizon} and Table~\ref{tab:main}.

\begin{figure}[t]
\centering
\includegraphics[width=0.86\textwidth]{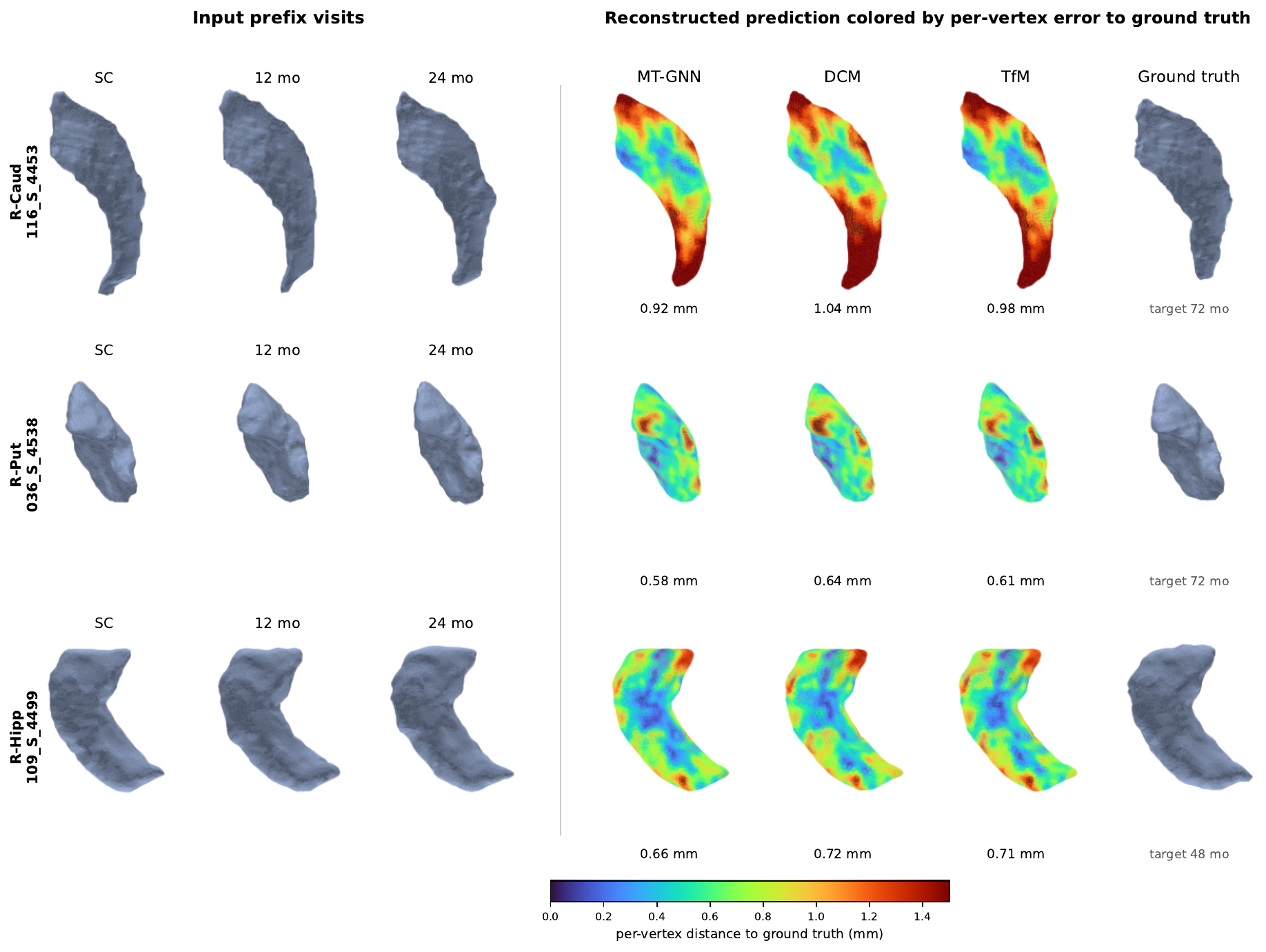}
\caption{\textbf{Qualitative forecasts on test subjects.}
For the right caudate, putamen and hippocampus (rows), the left block shows the three-visit causal
input prefix (screening, 12 and 24 months; shaded) and the right block the forecasts of MT-GNN, DCM
and TransforMesh at the target visit ($\Delta t{=}48$ for caudate and putamen, $\Delta t{=}24$ for
hippocampus) and the ground truth. Each prediction is rigidly aligned to the ground truth and colored
by per-vertex distance to it; the value below each panel is
the mean vertex error (MVE).}
\label{fig:qual}
\end{figure}

Figure~\ref{fig:qual} illustrates one example subject per structure. On the caudate, MT-GNN avoids the saturated error that DCM and TransforMesh leave along the tail; on the smoother putamen the three run close, with MT-GNN leaving the fewest hot spots; on the hippocampus, MT-GNN suppresses the error at the head that both baselines incur, and it attains the lowest MVE in every row.

\subsection{Ablations}
\label{sec:exp:ablation}
Table~\ref{tab:ablation} isolates each design choice.
The metric-space Ebin loss is the area-weighted metric on the space of Riemannian metrics, with the
affine-invariant SPD distance as its pointwise ingredient~\cite{ebin1970manifold,pennec2006riemannian}.
Training the identical model with this loss on $\hat g$ instead of the
MVE-through-ARAP objective drops the result from $-2.26\%$ to $-1.78\%$: supervising
through the reconstruction is consistently better, because the metric-space distance is only
loosely correlated with reconstructed vertex error. The gain is modest rather than dramatic, and
both objectives still beat the temporal mean---in this architecture the log-Euclidean base
already supplies a near-realizable metric, so the soft realizability constraint refines the prediction rather than rescuing it from collapse.

Adding the base shift improves the result from $-1.85\%$ (MT-GNN w/o base-shift) to $-2.26\%$ (MT-GNN)
(better on $11/14$ structures; patient-level paired $t$-test $p{=}4.0\times10^{-3}$), with the largest
gains on low-signal structures (e.g.\ right putamen $-0.29\%\rightarrow-1.91\%$). The
log-Euclidean base beats a Euclidean (arithmetic) mean of the input metrics ($-2.26\%$ vs.\ $-1.80\%$), and a
slope-shrinkage sweep is monotonically worse the more the per-subject slope is extrapolated,
supporting the mean-based design.

Adding a learned mean curvature head (MT-GNN+H) lowers the horizon-averaged mean error and is lower at
three of the four horizons (Table~\ref{tab:main}; at $\Delta t{=}36$ it is marginally worse than MT-GNN),
but the improvement over MT-GNN is small (mean $-0.13$ percentage points)
and reaches per-structure significance in only $4/14$ structures (Table~\ref{tab:pvalues}),
while never being significantly worse. Given this marginal, mostly-localized gain we retain the
simpler $g$-only MT-GNN as the primary model and report MT-GNN+H as a variant.

\begin{table}[htbp]
\caption{\textbf{Ablations}, single-setting on the causal test pairs (MVE
$\Delta\%$ vs.\ the temporal mean, as mean\,$\pm$\,SD across the 14 structures; each row changes one component of the full model). Every
variant still beats the temporal mean; training through the reconstruction, the base shift,
and the log-Euclidean base each help. Bold marks the retained model.}
\label{tab:ablation}
\centering
\setlength{\tabcolsep}{6pt}
\begin{tabular}{llc}
\toprule
Axis & Variant & $\Delta\%$ \\
\midrule
Training loss  & Ebin                        & $-1.78 \pm 1.22$ \\
               & \textbf{MVE}                & $\mathbf{-2.26 \pm 1.39}$ \\
\midrule
Base shift     & no shift                    & $-1.85 \pm 1.51$ \\
               & \textbf{MT-GNN}             & $\mathbf{-2.26 \pm 1.39}$ \\
\midrule
Mean geometry  & Euclidean base              & $-1.80 \pm 1.64$ \\
               & \textbf{Log-Euclidean base} & $\mathbf{-2.26 \pm 1.39}$ \\
\midrule
Curvature head & MT-GNN+H                     & $-2.39 \pm 1.25$ \\
               & \textbf{MT-GNN}             & $\mathbf{-2.26 \pm 1.39}$ \\
\bottomrule
\end{tabular}
\end{table}

%% file: sections/05_discussion.tex
\section{Discussion and Limitations}
\label{sec:discussion}

Per-visit subcortical change lies near the noise floor. The leading failure mode is thus amplifying
noise rather than underfitting. The trajectory extrapolation of DCM overshoots at long
horizons, and direct deformation models can spend capacity on visit noise. MT-GNN sidesteps both.
It predicts a small intrinsic residual on a geometry-aware mean, together with a population trend,
and it trains against the same reconstruction-constrained objective on which it is judged. The small
size of the curvature gain has a sharper geometric root than model capacity. We set that root out
next.

The predicted metric $\hat g$ fixes the edge lengths $\hat\ell$ through Eq.~\eqref{eq:edgelen}.
Once those lengths are set, the surfaces that bear the metric on our fixed connectivity make up the
isometric-realization set of a triangulated genus-zero mesh. For a generic mesh this set is
\emph{zero-dimensional}. The bar-and-joint framework is isostatic, since the free-vertex count
$3|\mathcal{V}|-|\mathcal{E}|$ equals the six degrees of freedom of a rigid motion, and a generic
framework is infinitesimally rigid \cite{gluck1975rigid}. No continuous bending is therefore
possible. What remains is only a finite number of branches, set apart by finite deformations rather
than by a continuum. An example of such a branch pair, sharing $\hat g$ but with opposite-sign $H$, is a planar cut with
either a depression or a cap. In classical surface
theory, mean curvature chooses among these branches and fixes the normal orientation that $\hat g$
leaves open. Adding $H$ fixes both principal curvatures, and hence the surface up to a rigid motion.
Our decoder settles that choice from the start. The ARAP solver begins at the temporal-mean surface
$\mathbf{S}^0$ and takes its rest directions from that same surface, following
\S\ref{sec:method:recon}. It thus falls into the one branch that holds $\mathbf{S}^0$, and it
converges to that branch's own solution, which is locally unique and infinitesimally rigid. The
branch and the orientation are just the knowledge that $H$ would supply, and the starting surface
has already fixed them. Mean curvature can therefore only sharpen the extrinsic shape within the
chosen branch. This is a second-order correction, and near the reproducibility floor it stays small.
Such is the pattern we see. $H$ helps only weakly and locally, reaching significance in four of the
fourteen structures, as Table~\ref{tab:pvalues} shows. The same argument shows when a curvature term
should weigh more. It should weigh more at long horizons, and under large, stereotyped atrophy, where
the change leaves the linear neighborhood of $\mathbf{S}^0$ and the branch is no longer easy to
choose. This fits the somewhat larger effect of the curvature head on the hippocampus and amygdala.

\paragraph{Limitations.}
We read realizability as an account of the gap between the metric-space and the reconstruction
objectives. It is not a theorem about exact projection. The ARAP solver is a robust local--global
method, not a Gauss--Codazzi enforcement. The gap stays small, since the log-Euclidean base is
already near-realizable, and we make no strong claim beyond this. The long-horizon experiments have few validation samples at the $36$- and $48$-month horizons. We therefore lead with the
mean and with the well-powered $12$- and $24$-month horizons, and we treat single long-horizon
per-structure cells as noisy. The load-bearing test is the paired per-structure $t$-test across
subjects, given in Table~\ref{tab:pvalues}. Our supervision sits on a $12$-month grid, although the
model runs continuously in $\Delta t$. Off-grid validation is left for further work. Finally, we
train one model per structure rather than a single multi-structure model.

%% file: sections/06_conclusion.tex
\section{Conclusion}
\label{sec:conclusion}

We presented MT-GNN, a continuous-time model of the future intrinsic first fundamental
form, for an arbitrary causal history and an arbitrary horizon. The prediction combines a
log-Euclidean mean base, a learned base shift, and a per-vertex subject residual. A differentiable
ARAP solver then decodes it into a surface. We tested the model on fourteen ADNI structures, with a
subject-disjoint held-out split and a shared metric and temporal-mean reference. MT-GNN and its
curvature variant MT-GNN+H are the two lowest-error predictors at every horizon. Their mean vertex
error falls $2.29\%$ and $2.48\%$ below the temporal mean, and they beat it on all fourteen
structures. Both stand ahead of geodesic regression and a mesh transformer, and their lead widens as
the horizon grows.

One finding backs this design. Training through the reconstruction can only ever decode to a valid
embedded mesh, and it consistently beats a metric-space loss on the metric tensors. Adding mean
curvature brings only a small gain, which reaches significance in four of the fourteen structures.
The reason is geometric. Once the metric is fixed, the surfaces that bear it form a zero-dimensional
set of separate branches \cite{gluck1975rigid}. Starting the reconstruction at the temporal mean
therefore chooses the branch and the orientation that curvature would otherwise give, and leaves it
a second-order refinement. We thus keep the metric alone as our main model. Further work spans a
unified multi-structure cortical and subcortical model, off-grid horizon validation, and clinical
progression endpoints.

%% file: sections/07_acknowledgements.tex
\section{Acknowledgments}
\label{sec:Acknowledgments}

This work was supported by NIH grant R01 MH131806. Data collection and sharing for this project was funded by the Alzheimer's Disease Neuroimaging Initiative (ADNI) (National Institutes of Health Grant U01 AG024904) and DOD ADNI (Department of Defense award number W81XWH-12-2-0012). 